\documentclass[10pt,twocolumn,letterpaper]{article}

\usepackage{cvpr}
\usepackage{times}
\usepackage{epsfig}
\usepackage{graphicx}
\usepackage{amsmath}
\usepackage{amssymb}
\usepackage[english]{babel}
\usepackage{algorithm}
\usepackage{multirow}
\usepackage{dsfont}
\usepackage{adjustbox}
\usepackage[font=Large]{subcaption}
\usepackage[font=small,skip=0pt]{caption}
\usepackage[noend]{algpseudocode}
\usepackage{pifont}
\usepackage{bbding}
\usepackage{amsthm}
\usepackage{float}

\newtheorem{theorem*}{Theorem}

\usepackage{authblk}
\usepackage[pagebackref=true,breaklinks=true,letterpaper=true,colorlinks,bookmarks=false]{hyperref}

\cvprfinalcopy 

\def\cvprPaperID{22} 

\ifcvprfinal\fi
\begin{document}
\newcommand{\STAB}[1]{\begin{tabular}{@{}c@{}}#1\end{tabular}}

\pagenumbering{gobble}
\title{Efficient Semantic Segmentation using Gradual Grouping}

\author[1]{Nikitha Vallurupalli}
\author[1]{Sriharsha Annamaneni}
\author[1]{Girish Varma}
\author[1]{\\ C V Jawahar}
\author[2]{Manu Mathew}
\author[2]{Soyeb Nagori}

\author[.]{\\{\small \tt nikitha.vallurupalli@research.iiit.ac.in, sriharsha0806@gmail.com, girish.varma@iiit.ac.in, jawahar@iiit.ac.in, mathew.manu@ti.com, soyeb@ti.com}}

\affil[1]{\small Center for Visual Information Technology,  Kohli Center on Intelligent Systems, IIIT-Hyderabad, India}
\affil[2]{\small Texas Instruments, Bangalore, India}

\maketitle
\begin{abstract}
Deep CNNs for semantic segmentation have high memory and run time requirements. Various approaches have been proposed to make CNNs efficient like grouped, shuffled, depth-wise separable convolutions. We study the effectiveness of these techniques on a real-time semantic segmentation architecture like ERFNet for improving runtime by over 5X. We apply these techniques to CNN layers partially or fully and evaluate the testing accuracies on Cityscapes dataset. We obtain accuracy vs parameters/FLOPs trade offs, giving accuracy scores for models that can run under specified runtime budgets.

We further propose a novel training procedure which starts out with a dense convolution but gradually evolves towards a grouped convolution. We show that our proposed training method and efficient architecture design can improve accuracies by over 8\%  with depthwise separable convolutions applied on the encoder of ERFNet and attaching a light weight decoder. This results in a model which has a 5X improvement in FLOPs while only suffering a 4\% degradation in accuracy with respect to ERFNet. 
\end{abstract}

\section{Introduction}
\label{sec:intro}
Semantic segmentation is a critical computer vision component of autonomous navigation and robotic systems. It involves dense and high dimensional prediction of a label for every pixel of an input image. In real world systems, it also needs to be done on a video stream at high frames per second, in a power efficient manner. Also there is major challenge of safety in systems such as autonomous navigation. Hence for models to be practically applicable, it is essential that they have to be compact, fast as well as achieve high prediction accuracies. 

Deep CNN based models have brought forward a giant leap in prediction accuracies in semantic segmentation \cite{girshick2014rich, long2015fully}. However they are computationally expensive and it is not clear if high accuracy models can be fitted in the resource constraints set by applications. Hence it has become one of the major challenges in deep learning, to make these models efficient while maintaining prediction accuracies. In the last few years, a new area commonly known as Model Compression has emerged which aims to address this challenge.

Initial attempts at model compression were inspired by matrix compression techniques. DNNs essentially consists of weight matrices and the obvious ways to make computations on matrices fast is by making them sparse. Such methods are commonly known as pruning techniques\cite{jaderberg2014speeding, lebedev2014speeding}. Another approach is quantization which is to round the weight matrices (typically floating point arrays) to integer arrays with lower precision. These methods have significantly reduced model sizes, however accuracy degrades at high compression rates \cite{han2015deep}, which makes training small models better. All these methods also require several phases of retraining so that prediction accuracies can be recovered after pruning/quantization.

\begin{figure*}[!tbh]
\centering
  
  \includegraphics[width=\textwidth]{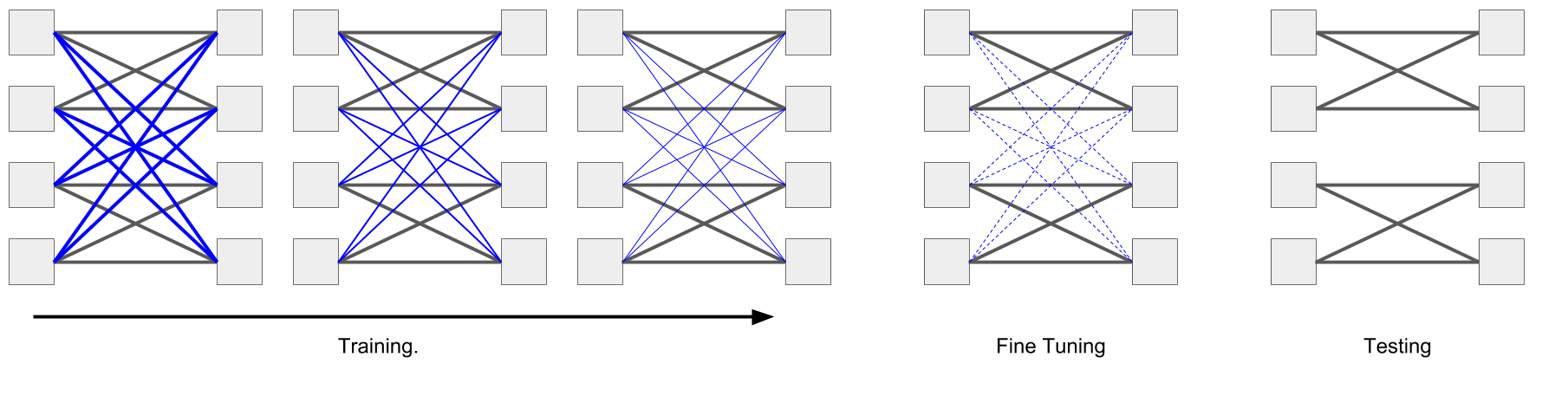}
  
\caption{Our proposed training procedure for obtaining improved accuracies in grouped convolution based architectures. Crucial observation is that grouped convolution can be thought of as a dense convolution with many weights being 0 (the blue edges). Note that here each edge represents a convolutional filter of $w\times w$. In our method, we start with a dense convolution and multiply the blue edges by a parameter $\alpha$. We decrease $\alpha$ gradually during training time from $1$ and by the end of the training it becomes $0$. We also have a fine tuning phase where $\alpha$ remains $0$. Finally at test time, the convolutions can be implemented as a grouped convolution which gives better efficiency. Since the optimization that is happening at training time is in the higher dimensional space of dense convolutions, we can obtain better accuracies than traditional training for grouped convolutions.}
\label{fig:grad_group_summ}
\end{figure*}

A more recent approach to model compression is to design the architecture with specific insights about the information flow required to give accurate predictions. Efficient layer designs started with GoogLeNet \cite{googlenet}, who proposed to reduce the input channels to 3x3 convolutions. Xception \cite{xception} took it further by using depth-wise 3x3 separable convolutions. Grouped convolutions \cite{xie2017aggregated} proposed a simple way of having structured sparsity in convolutions. Very recently shuffled convolutions \cite{zhang2017shufflenet} have been proposed which improves upon grouped convolutions by doing a shuffle operation after the grouping. 

We study the effectiveness of depthwise separable, grouped and shuffled convolutions on a realtime and efficient semantic segmentation model ERFNet \cite{romera2018erfnet}. We replace the modules in ERFNet with depthwise separable modules. We experiment with different group number and shuffle operations. We observe that these methods can reduce the FLOPs significantly but incurs as much as 10\% degradation in accuracies (see Section \ref{sec:DGS*}).

We propose a novel training framework for grouped convolutions, called gradual grouping (see Section \ref{sec:res_grad}). In this training method, we gradually evolve a dense convolution toward a grouped convolution. This allows the gradient descent to happen at a higher dimensional model space initially and gradually evolving towards a lower dimensional subspace of grouped convolutions. Our approach is inspired by lifting methods in linear programming where better optimization can be done in a higher dimensional representation.

We use our training procedure to obtain a model which is only 5.77 GFLOPs (5X improvement over ERFNet which is 27.7 GFLOPs) with 68\% accuracy (4\% reduction over ERFNet which gives 72\%) (see Section \ref{sec:res_grad}).  We also find models which gives 1.5X, 2X reduction in FLOPs with only 0\%, 2\% reduction in accuracies respectively (see Section \ref{sec:select}). 

\section{Related Works}
\label{sec:related-works}
Our work mainly focuses on designing efficient semantic segmentation architectures and training methods that enhance the computational efficiency. A substantial amount of work has been done in comparing our proposed models with the existing realtime semantic segmentation architectures. We also discuss about the wide variety of model compression techniques that have been proposed in the recent years.
\subsection{Realtime Semantic Segmentation}
Different classes of deep learning based semantic segmentation architectures have been proposed. Most of the models follow the fully convolutional networks (FCNs) \cite{long2015fully} approach. Early work in designing convolutional neural networks architecture for semantic segmentation concentrated on accuracy  (weighted IOU). Most of the semantic segmentation models follow an Encoder-Decoder type of architecture. In the encoder part of these networks, the feature extractors are powerful object detectors like ResNet, ResNext, etc. PSPNet \cite{zhao2017pyramid} achieves accuracies above 80\%. However PSPNet \cite{zhao2017pyramid}, runs at more than 100 GFLOPs. Our work is more focused on obtaining models with $< 20$ GFLOPs.

More recent works that focus on realtime efficient segmentation are ERFNet \cite{romera2018erfnet}, ENet \cite{paszke2016enet}, ICNet \cite{zhao2017icnet}, SegNet basic  \cite{badrinarayanan2017segnet} and Clockwork FCNs \cite{shelhamer2016clockwork}. However, all of them propose architectural modifications. Our work is more focused on using efficient CNN modules and better training procedures by keeping the macro architecture the same. In this paper, we adopt the ERFNet macro architecture and experiment with depthwise separable, grouped and shuffled convolutions applied to it. 

\subsection{Model Compression}
Model compression refers to the broad set of techniques that makes models compact. Initial methods proposed include pruning and quantization techniques \cite{han2015deep}. There have been works which apply these techniques for semantic segmentation \cite{8014780}. However in \cite{8014780}, the focus was on a coarse segmentation on only a few classes, while we are attempting to build efficient models for the Cityscapes benchmark \cite{cordts2016cityscapes} with all the classes.

Newer approaches to model compression, involves designing efficient CNN Modules. GoogleNet \cite{googlenet} proposed inception modules which decrease the channels to expensive 3x3 convolutions. Xception and MobileNet took this further to make 3x3 convolutions completely depthwise separable and sparse. ResNext \cite{xie2017aggregated} employed grouped convolutions to get efficient models. Shufflenet \cite{zhang2017shufflenet} improved upon grouped convolutions further by adding a shuffling layer which helped in better mixing of information across channels. However these works have mostly focused on the classification benchmarks like Imagenet \cite{deng2009imagenet}. We focus on applying some of these methods to the task of semantic segmentation, on Cityscapes dataset. We also propose training methods that work well with grouped convolutions.

\subsection{Architecture Search}
Very recently there have been works on architecture search \cite{pham2018efficient, zoph2017learning, zoph2016neural} where a separate machine learning algorithm is used to drive a heuristic search procedure to pick an efficient architecture. However these methods require large server farms, and has not yet proved its utility for a dense prediction task like semantic segmentation. CondenseNet \cite{huang2017condensenet} proposes a simple architecture search procedure integrated with the training of the base network focusing on classification benchmarks. In our work, the architecture is fixed before hand unlike \cite{huang2017condensenet}. However we propose novel training algorithms which give better accuracies for grouped convolutions.

\section{Approach}
\label{sec:approach}
We propose to obtain extremely efficient semantic segmentation architectures by devising specialized training techniques for grouped convolutions. We do our experiments on ERFNet \cite{romera2018erfnet} which is already an efficient and realtime model. In Section \ref{sec:eff_cnns}, we describe the efficient CNN layer designs that we use. Finally in Section \ref{sec:grad_group}, we describe our novel training procedure.

  

\subsection{ERFNet}\label{sec:erfnet}
\begin{figure*}[!tbh]
\centering
  \includegraphics[scale=0.50]{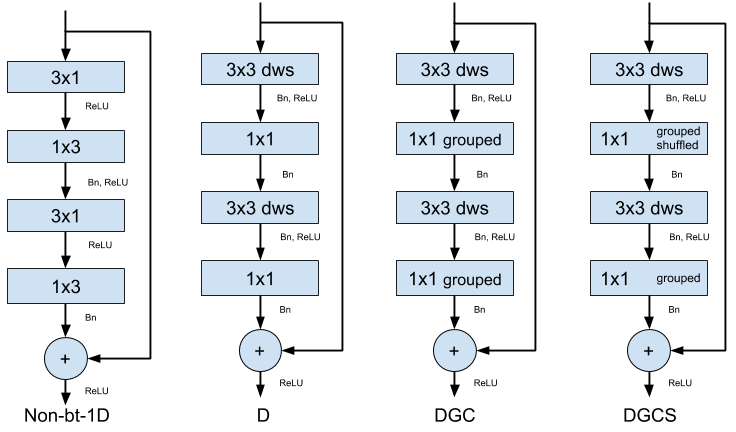}
  
\caption{Different types of layers used in the proposed architecture.
Non-bt-1D is the non-bottleneck layer used in ERFNet, D, DGC and and DGCS are our proposed layer architectures}
\label{fig:cnn_modules}
\end{figure*}

ERFNet \cite{romera2018erfnet} proposes an efficient convolutional block as the core of the architecture which achieves state-of-the-art accuracy on the Cityscape dataset at real time. It was proposed as an improvement over the ENet \cite{paszke2016enet}, which is highly efficient (runs at $<2$GFLOPS), but has low accuracies (57\% IOUs).  ERFNet model obtains 70\% accuracy at $27.7$ GFLOPs. The ERFNet architecture has an encoder and decoder parts. ERFNet encoder pretrained on ImageNet \cite{deng2009imagenet} achieves better accuracy on the Cityscapes dataset \cite{cordts2016cityscapes} than ERFNet trained from scratch.
It consists of convolutional modules which they call Non-bottleneck-1D (see Figure \ref{fig:cnn_modules}), which uses spatial separability by using an $1\times 3$ and $3\times 1$ convolutions. 

\subsection{Efficient CNNs}\label{sec:eff_cnns}

\subsubsection{Depthwise separable convolutions}

Depthwise separable convolutions \cite{6619199,xception}, comprise of a depthwise convolution performed over each channel of an input layer and followed by a $1 \times 1$ convolution. $1 \times 1$  convolution is called pointwise convolution that takes the output channels from previous step and then combines them into an output layer. Compared to normal convolution, there is a reduction in the number of parameters which decreases the computation required and model size as well. Using depthwise separable convolution we can reduce the computational cost by 1/C$_{out}$ + 1/$K^2$ where C$_{out}$ is the channel output size and K is the filter size of the convolution layer \cite{howard2017mobilenets}.

    



The reduction in the parameters make separable convolutions quite efficient with improved runtime performance. They also have the added benefit of reducing over-fitting to an extent, because of the fewer parameters. Depth wise separable convolutions are used in models like MobileNet \cite{howard2017mobilenets}, Xception \cite{xception} and ResNeXt \cite{xie2017aggregated}.

In our proposed convolutional module (see D, in Figure \ref{fig:cnn_modules}), we use depthwise separable convolutions in place of spatially separable convolutions used in Non-bt-1D layer.


\begin{figure}\label{fig:grouping}
\includegraphics[scale=0.25]{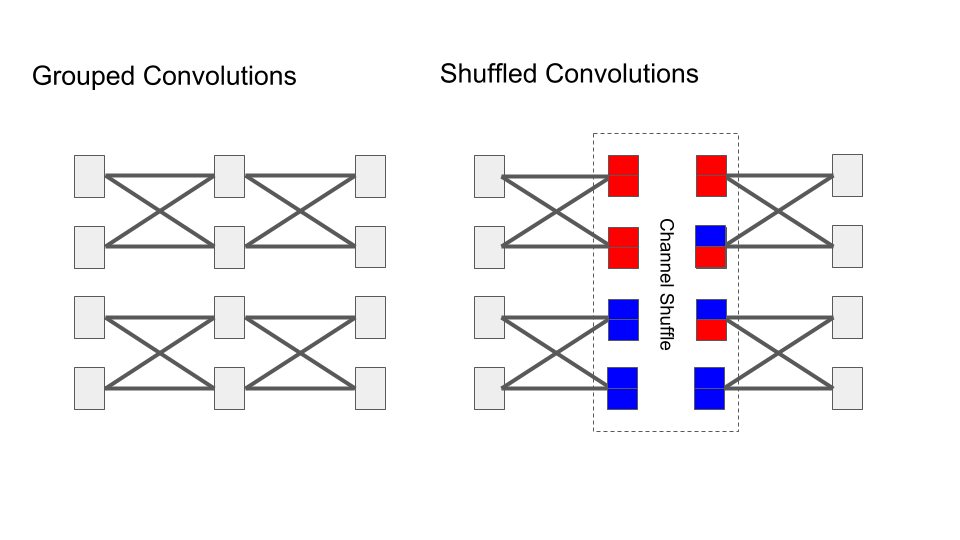}
\caption{Grouping and Shuffling}
\end{figure}
\subsubsection{Grouped Convolutions}
Grouped convolution is another way of building structured sparse convolutions. Such a convolution with groups parameter $g$, decreases the parameter and FLOPs of the layer by a factor $g$. Grouped  
convolutions also help in data bandwidth reduction \cite{8014780}, was first implemented by AlexNet \cite{krizhevsky2012imagenet}. It has been used for efficient layer design in ResNext \cite{xie2017aggregated}. 
In our proposed convolutional module (see DGC, in Figure \ref{fig:cnn_modules}) groups are applied to the 1x1 pointwise convolutions which otherwise consume huge number of parameters in the network architecture. 
\begin{table}[!htbp]
 \centering
 \caption{Network Architecture with proposed Layers-[D*]}
 \label{tab:arch1}
 \resizebox{\columnwidth}{!}{
 \begin{tabular}{|l|c|c|c|c|}
 \cline{1-5}
 \hline 
 	& \textbf{Layer} & \textbf{Type} & \textbf{out-chann} & \textbf{out-Res} \\[1.5ex] \cline{1-5}
     \hline
     \hline
     \multirow{12}{*}{\STAB{\rotatebox[origin=c]{90}{ENCODER}}}
  	& 1	&  \textbf{Downsampler block} & 16 & 512x256 \\ \cline{2-5}
     & 2 & \textbf{Downsampler block} & 64 & 256x128 \\
     & 3-7 & \textbf{5 x Conv-module} & 64 & 256x128 \\ \cline{2-5}
     & 8 & \textbf{Downsampler block} & 128 & 128x64 \\
     & 9 & \textbf{Conv-module}(dilated 2) & 128 & 128x64 \\
     & 10 & \textbf{Conv-module}(dilated 4) & 128 & 128x64 \\
     & 11 & \textbf{Conv-module}(dilated 8) & 128 & 128x64 \\
     & 12 & \textbf{Conv-module}(dilated 16) & 128 & 128x64\\
     & 13 & \textbf{Conv-module}(dilated 2) & 128 & 128x64\\
     & 14 & \textbf{Conv-module}(dilated 4) & 128 & 128x64\\
     & 15 & \textbf{Conv-module}(dilated 8) & 128 & 128x64\\
     & 16 & \textbf{Conv-module}(dilated 16) & 128 & 128x64 \\  \cline{1-5}
     \multirow{3}{*}{\STAB{\rotatebox[origin=r]{90}{DECODER}}}
     & 17 & \textbf{Deconvolution}(upsampling) & 64 & 256x128 \\
     & 18-19 & \textbf{2 x Non-bt-1D} &  64 & 256x128 \\ \cline{2-5}
     & 20 & \textbf{Deconvolution}(upsampling) & 16 & 512x256 \\ 
     & 21-22 & \textbf{2 x Non-bt-1D} & 16 & 512x256 \\ \cline{2-5}
     & 23 & \textbf{Deconvolution}(upsampling) & C & 1024x512 \\ \cline{1-5}  
    
 \end{tabular}}
 \end{table}
\subsubsection{Channel Shuffling}
If multiple group convolutions are stacked together, outputs from a certain channel are only derived from a small fraction of input channels. It is clear that outputs from a certain group only relate to the inputs within the group. This property blocks information flow between channel groups and weakens representation. 

By doing grouping in 1x1 point wise convolution, we could reduce the number of parameters consumed, but the information flow across groups is blocked. If we allow group convolution to obtain input data from different groups, the input and output channels will be fully related. Specifically, for the feature map generated from the previous group layer, we can first divide the channels in each group into several subgroups, then feed each group in the next layer with different subgroups. This can be efficiently implemented by a channel shuffle operation \cite{zhang2017shufflenet}. Channel shuffling operation enables cross-group information flow for multiple group convolutions. In our proposed layer architecture (see DGCS, in Figure \ref{fig:cnn_modules}), we do a channel shuffle operation after a grouped 1x1 point wise convolution before passing information to the next convolution block.

\subsection{Gradual Training of Grouped Convolutions}\label{sec:grad_group}

We propose a novel training procedure with special focus on grouped convolutions, which can improve the accuracies (see Figure \ref{fig:grad_group_summ}). We first observe that grouped convolution can be thought of as a dense convolutions with certain weights zeroed out. Hence the space of a grouped convolutions is nothing but a linear subspace of dense convolutions. Traditional training procedures starts out with a grouped convolution model, and hence the gradient descent optimization will only happen in the low dimensional subspace of grouped convolutions. It is a well known result in linear programming lifting that optimization in a higher dimensional space can often lead to convergence towards better minima.

We propose a training procedure where the train time optimization happens in the higher dimensional space of dense convolutions and gradually evolves toward a grouped convolutions. In this proposed training procedure, the network starts out as a model with no groups which is equivalent to saying that the total number of groups is equal to one and gradually evolves to model with number of groups equal to 2. In this process of training, the dense connections which we had initially gradually reduce to sparse group connections. At the time of test, the model has connections only within groups and can be implemented as grouped convolution. This reduces the number of FLOP's significantly at the time of validation and testing and we see that there is no accuracy drop, as the group size increases which also signifies that accuracy is not degrading as the model size decreases.

\section{Experimental Details}
\label{sec: experiments}
\subsection{Network Configuration}\label{sec:network_config}
The basic segmentation architecture is inspired from ERF Net, considering the balance between FLOP count and accuracy. 
In the proposed architecture (see Table \ref{tab:arch1}), we change the encoder completely by replacing each Non-bt-1D layer with our proposed convolutional layers and this architecture is named - [D*]. When the Conv-module (see Table \ref{tab:arch1}) is configured with DGC architecture (see Fig \ref{fig:cnn_modules}), it is named [DG(C)*] where C is the value of number of groups.

If Conv-module has DGCS architecture (see Fig \ref{fig:cnn_modules}), then the network is named [DG(C)S*] where C indicates the number of groups and S indicates that a shuffle operation is being done.
More detailed architecture of this network describing all the layers is seen in Table \ref{tab:arch}. All the results are reported and compared using this nomenclature. 
\begin{table}[!htbp]
 \centering
 \caption{Selective Application of proposed Layers in the Network Architecture -[D]}
 \label{tab:arch}
 \resizebox{\columnwidth}{!}{
 \begin{tabular}{|l|c|c|c|c|}
 \cline{1-5}
 \hline 
 	& \textbf{Layer} & \textbf{Type} & \textbf{out-chann} & \textbf{out-Res} \\[1.5ex] \cline{1-5}
     \hline
     \hline
     \multirow{12}{*}{\STAB{\rotatebox[origin=c]{90}{ENCODER}}}
  	& 1	&  \textbf{Downsampler block} & 16 & 512x256 \\ \cline{2-5}
     & 2 & \textbf{Downsampler block} & 64 & 256x128 \\
     & 3-5 & \textbf{3 x Non-bt-1D} & 128 & 128x64 \\
     & 5-7 & \textbf{2 x Conv-module} & 64 & 256x128 \\ \cline{2-5}
     & 8 & \textbf{Downsampler block} & 128 & 128x64 \\
     & 9 & \textbf{Non-bt-1D}(dilated 2) & 128 & 128x64 \\
     & 10 & \textbf{Non-bt-1D}(dilated 4) & 128 & 128x64 \\
     & 11 & \textbf{Non-bt-1D}(dilated 8) & 128 & 128x64 \\
     & 12 & \textbf{Non-bt-1D}(dilated 16) & 128 & 128x64\\
     & 13 & \textbf{Conv-module}(dilated 2) & 128 & 128x64\\
     & 14 & \textbf{Conv-module}(dilated 4) & 128 & 128x64\\
     & 15 & \textbf{Conv-module}(dilated 8) & 128 & 128x64\\
     & 16 & \textbf{Conv-module}(dilated 16) & 128 & 128x64 \\  \cline{1-5}
     \multirow{3}{*}{\STAB{\rotatebox[origin=r]{90}{DECODER}}}
     & 17 & \textbf{Deconvolution}(upsampling) & 64 & 256x128 \\
     & 18-19 & \textbf{2 x Non-bt-1D} &  64 & 256x128 \\ \cline{2-5}
     & 20 & \textbf{Deconvolution}(upsampling) & 16 & 512x256 \\ 
     & 21-22 & \textbf{2 x Non-bt-1D} & 16 & 512x256 \\ \cline{2-5}
     & 23 & \textbf{Deconvolution}(upsampling) & C & 1024x512 \\ \cline{1-5}

 \end{tabular}}
 \end{table}

We come up with another Network Architecture (see Table \ref{tab:arch}) based on Selective Application of proposed layers. As discussed earlier, in D* Networks, all the layers are changed which results in an accuracy drop. From conventional model compression techniques like pruning and quantization, we adapt the thought to apply compression techniques only to the later layers in the network. In this network architecture, namely-[D] (see Table \ref{tab:arch}), we apply our proposed Conv-module layers selectively, leaving few initial layers after Downsampler block. When the Conv-module layer is configured with DGC (see Fig \ref{fig:cnn_modules}), network is named [DG(C)] where is C is the number of groups. When a channel shuffle layer is used in between grouped convolutions, we call the network [DG(C)S]. Dilation is used in the layers to gather more context information{\cite{romera2018erfnet}}. Downsampler block architecture(see Table \ref{tab:arch}) includes deconvolution layers with stride 2, which is the same as  transposed convolutions. Deconvolutions simplify memory and computation requirements, unlike max-unpooling operation as sharing the pooling indices from the encoder is not required. We use a  small decoder to reduce the number of parameters, whose purpose is to upsample the encoder’s output by fine-tuning the details{\cite{romera2018erfnet}}. 

\subsection{Experimental Setup}

We use Cityscapes Dataset {\cite{cordts2016cityscapes}} in all our experiments. It is a 
challenging dataset with 19 labeled classes. It contains a train set of 2975 images, a validation set of 500 images and a test set of 1525 images. All the models are trained only using the train set. To access the performance of the architecture we use Intersection over Union (IoU) scores as accuracy metric. We report meanIoU, which is the validation accuracy on all the 19 classes. During the time of test and validation, image is sub sampled by a factor of 2 to report meanIoU.

 All the proposed network architectures (see Sec \ref{sec:network_config}), namely D*, DG*, D(G)S* (see Table \ref{tab:arch1}), D, DG and D(G)S (see Table \ref{tab:arch}) are trained from scratch on cityscapes dataset. All the experiments are done in pytorch with CUDA 9.0 and CUDNN back ends. Adam optimizer \cite{kingma2014adam} of stochastic gradient decent is used for training. Training is done with a batch size that is inversely proportional to the size of the proposed compressed models. We start with a learning rate of $5*e^{-04}$, and a learning rate scheduler is used to decrease the learning rate, so that the convergence is accelerated.
\subsection{Training Procedure for gradual grouping}
In the proposed training procedure, a D (see Table \ref{tab:arch}) , D* encoder model (see Table \ref{tab:arch1}) evolves to a DG, DG* encoder model (see Sec \ref{sec:network_config}) respectively. G is the targeted number of groups. In this process, the encoder and decoder are trained in two phases. The meanIoU value is calculated only after the model settles to a DG, DG* model respectively. The training is done using a controllable parameter alpha. As the value of alpha changes with increasing number of epochs, the connections become sparse (see Fig \ref{fig:grad_group_summ}). When the value of alpha is 1, it is a D, D* (see Sec \ref{sec:network_config}) model respectively.

In the initial epochs, alpha gradually decrements from 1 to 0. When alpha value becomes zero, it is DG, DG* model respectively (see Sec \ref{sec:network_config}). D, D* model evolves to a DG, DG* encoder model where G is the targeted number of groups. In the last few epochs, the model is fine-tuned keeping alpha value zero and the model converges. As the grouping techniques are applied only in the encoder,the controllable parameter alpha is used only in training the encoder.

Using these pretrained encoder weights, the encoder decoder architecture is trained. 
We remove the last layer from the encoder and attach the decoder in order  to train the full network. Now, since the encoder model uses pretrained gradual grouping weights, the encoder is well initialized but decoder weights are not trained. When the encoder is trained again along with decoder, the initialization gained through gradual grouping is lost. To overcome this, we almost freeze the encoder which is equivalent to giving a very less learning rate of $5*e^{-20}$ to the encoder for few initial epochs. Whereas the decoder will have a learning rate of $5*e^{-04}$ in the initial epochs. In the later epochs, the encoder and decoder start training together with the same learning rate of $5*e^{-04}$. 
The proposed novel training procedure can be easily implemented, which specifically targets grouped convolutions.


\section{Results}

Our main result is to obtain a semantic segmentation model with  5.8 GFLOPs running time with IOU scores of 68\%. Our baseline model is ERFNet which is a realtime semantic segmentation model and our method gives a 5X reduction in FLOPs with only 4\% degradation in accuracy. All the accuracies reported is by the same procedure, where ground truths are sub sampled to half the resolution, and compared with the predictions \footnote{We do evaluation with ground truths sub sampled by factor 2. This was done due the limitation of the number of GPU's available.}.

Our approach is to first apply  depthwise separable layers along with grouping and shuffling operations on the entire ERFNet architecture to reduce the GFLOPs significantly (see Section \ref{sec:DGS*}). However this process degrades the accuracy also significantly, by over 10\%. We propose a special training procedure called gradual grouping where the accuracy degradation is reduced resulting in a compact network with 6X reduction in runtime and 66\% IOUs (see Section \ref{sec:res_grad}). We also apply grouping and shuffling operations selectively to layers of the ERFNet architecture, resulting in models with accuracies similar to the baseline with better FLOPs/parameter tradeoffs (see Section \ref{sec:select}).

\begin{figure*}[!tbh]\label{fig:perf_tradeoffs}
\includegraphics[width=\textwidth]{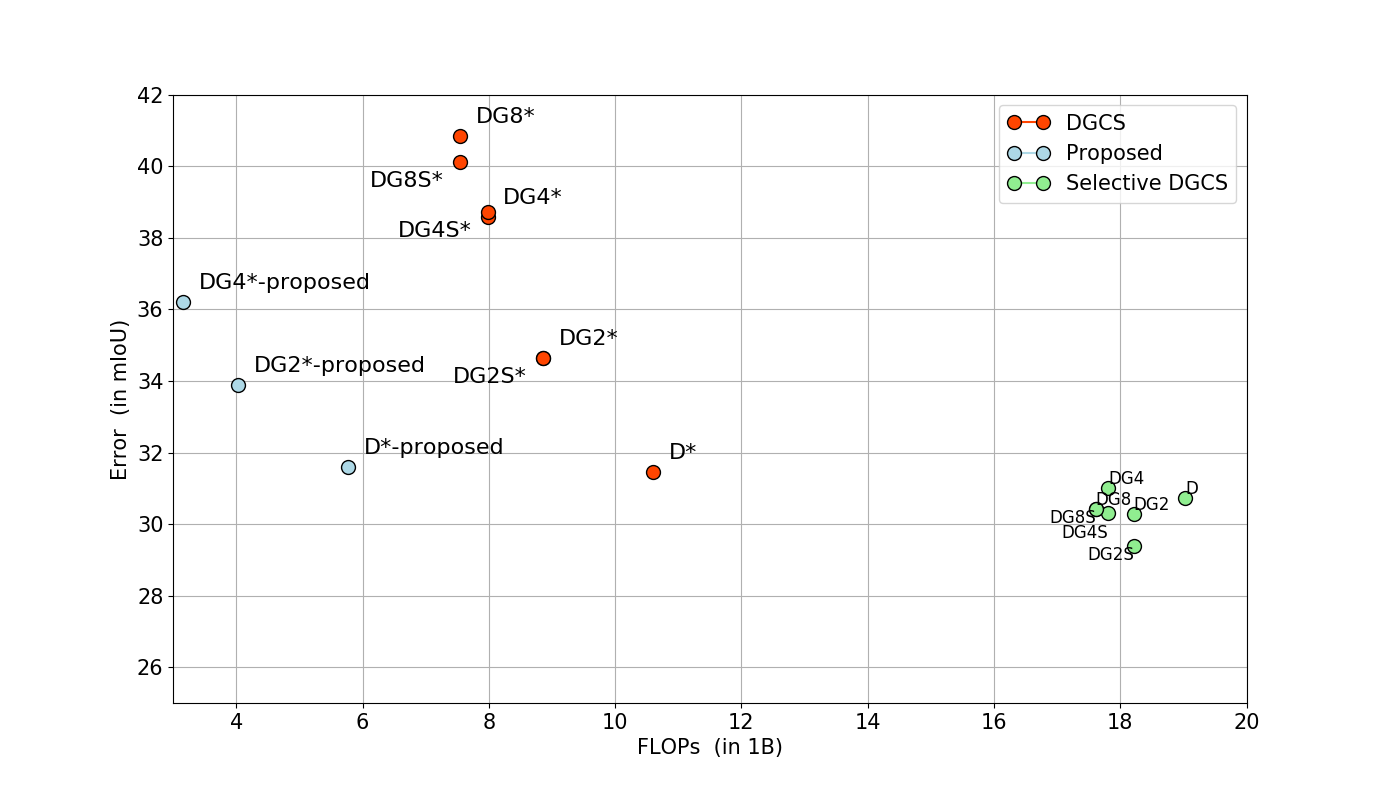}
\caption{Performance trade off graph for all the models studied. Note that the blue points representing models trained by gradual grouping gives the best performance tradeoffs. Also the selective application of groups (green points) hardly degrades the accuracy while still giving a reasonable reduction in GFLOP of 1.5X over the baseline ERFNet which runs at 27.7 GFLOPs.}
\vspace{-1em}
\end{figure*}

\subsection{Comparison with Depthwise separable, Groups and Shuffle Layers} \label{sec:DGS*}

\begin{table}[!htbp]
\centering
\caption{Depthwise Separable Convolution, Groups and Shuffle on ERFNet Architecture}
\label{tab:dgs_star}
\begin{tabular}{|l|r|r|r|}
\cline{1-4}
\textbf{Models}      & \textbf{IOU}   & \textbf{Params}  & \textbf{GFLOPs} \\ \cline{1-4}
\hline
\hline
ERFNet & 70.45& 2038448& 27.705\\ \cline{1-4}
D* &68.55& 547120 & 10.597\\ \cline{1-4}
DG2* & 65.35 & 395568  & 8.852  \\ \cline{1-4}
DG4* & 61.42 & 319792 & 7.980  \\ \cline{1-4}
DG8* & 59.15 & 281904 & 7.543 \\ \cline{1-4}
DG2S* & 65.36 & 395568 & 8.852  \\ \cline{1-4}
DG4S* &61.27& 319792 & 7.980 \\ \cline{1-4}
DG8S* &59.89 & 281904 & 7.543 \\ \cline{1-4}
\end{tabular}
\end{table}

We study the effect of various efficient CNN designs on the ERFNet 
architecture for semantic segmentation. First, we replace the spatially 
separable modules in ERFNet by an equivalent depthwise separable 
convolution modules (see Figure \ref{fig:cnn_modules}), in most of the 
layers (see Table \ref{tab:arch}) resulting in the model D*. As can be 
seen in Table \ref{tab:dgs_star}, this reduces the FLOPs by almost 3X with an accuracy degradation of 2\%, which is quite good. 

We further use grouped convolutions for decreasing the FLOPs. Since most of the parameters in the depthwise separable convolutions are in the 1x1 point-wise convolutions, we use the grouping only on the 1x1 convolutions. This results in the models DG2*, DG4*, DG8* with group sizes 2, 4 and 8 respectively. These models are sufficiently compressed in terms of FLOPs with the DG8* model being only 7.5 GFLOPs. However the accuracy degradation is over 10\% which is likely to be unacceptable.

Shuffled convolutions was proposed to improve the accuracies of grouped convolutions. They have the same parameters and FLOPs as grouped convolutions, since shuffle operation is essentially just a rearranging of the channels. The models DGCS* (where C=2,4,8) are also mentioned in Table \ref{tab:dgs_star}. But we observe, that shuffling operation is not affecting the accuracy in our case.

\subsection{Training using Gradual Grouping}\label{sec:res_grad}
As seen in Section \ref{sec:DGS*}, grouped convolutions can significantly reduce the FLOPs, but incur an unacceptable reduction in accuracy. We propose a novel training procedure which starts with a full dense model and gradually evolves toward a model with larger group size. 

	We further try to reduce the number of FLOPs by attaching a light weight decoder to the existing encoder model which we proposed. We compress the decoder further by changing the 3x3 deconvolution (upsampling) operation to 1x1 upsampling and also changing the Non-bt-1D layers to Conv-module layers (see Table \ref{tab:arch1}). We train our proposed encoder using gradual grouping on Imagenet dataset and then attach the proposed light weight decoder to it.
    
	The results of the gradual grouping training is given in Table \ref{tab:grad_group}. As it can be seen, the FLOPs vs accuracy trade-off has decreased significantly with our proposed method on the modified architecture. Specifically the DGC* models when trained with gradual grouping gives a significant improvement in accuracy over the normal training. Our proposed model is giving accuracies of 68\% (4\% reduction from baseline ERFNet) while having only 5.77 GFLOPs (5X reduction in FLOPs).

\begin{table}[ht]
\centering
\caption{Gradual Training of Grouped Convolutions. As can be seen, our proposed models are having FLOPs ranging from 5.77 GFLOPs to 3.15 GFLOPs while the best accuracy is around 68\%. Improvement in accuracy is seen due to gradual training from the traditional training method (reported in Table \ref{tab:dgs_star})}
\label{tab:grad_group}
\begin{tabular}{|l|r|r|r|}
\cline{1-4}
\textbf{Models} & \textbf{IOU} & \textbf{Params}  & \textbf{GFLOPs}\\ \cline{1-4}
\hline
\hline
ERFNet-pretrained & 72.10 & 2038448 & 27.705  \\ \cline{1-4}
D*-proposed & 68.39 & 431312 & 5.773  \\ \cline{1-4}
DG2*-proposed & 66.10 & 279760 & 4.029  \\ \cline{1-4}
DG4*-proposed & 63.80 & 203984 & 3.156  \\ \cline{1-4}
\end{tabular}
\vspace{-1em}
\end{table}

\subsection{Selective Application of Groups} \label{sec:select}
We further experiment with models where we selectively replace some 1x1 convolutions in the D* model with grouped and shuffled convolutions. The specific layers that are replaced is given in Table \ref{tab:arch}. The results of these experiments are given in Table \ref{tab:dgs}. The models DGC's refer to models were groups = C is used for selected 1x1 convolutions. We also did experiments with the corresponding shuffle convolution versions also. We observe that the model DG2S without decoder incurs only a 1\% reduction in accuracy while giving a 2X improvement in FLOPs. Also the model DG2S gives a 1.5X improvement in FLOPs without incurring any loss in accuracy.

\begin{table}[h]
\centering
\caption{Selective Application of Depthwise separable convolutions, Grouping and Shuffling}
\label{tab:dgs}
\begin{tabular}{|c|c|c|c|}
\hline
\textbf{Models}      & \textbf{IOU}   & \textbf{Params}  & \textbf{GFlops}  \\ \cline{1-4}
\hline
\hline
D   &  69.26 & 1291648 & 19.025 \\ \cline{1-4}
DG2 &  69.71 & 1238960 & 18.998 \\ \cline{1-4}
DG4 &  68.98 & 1202096 & 18.595 \\ \cline{1-4}
DG8 &  69.57 & 1183664 & 18.394 \\ \cline{1-4}
DG2S &  70.62 & 1238960 & 18.998 \\ \cline{1-4}
DG4S & 69.59 & 1202096 & 18.595 \\ \cline{1-4}
DG8S & 69.57 & 1183664 & 18.394 \\ \cline{1-4}
\end{tabular}
\vspace{-1em}
\end{table}



\begin{figure*}
\label{Qualitative_examples}
\begin{subfigure}{0.5\textwidth}
	\includegraphics[scale=1.5, width=\textwidth]{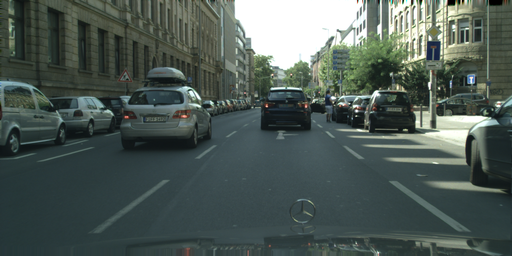}
\end{subfigure}
\begin{subfigure}{.5\textwidth}
	\includegraphics[scale=1.5, width=\textwidth]{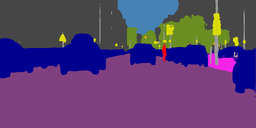}
\end{subfigure}
\begin{subfigure}{.5\textwidth}
	\includegraphics[scale=1.5, width=\textwidth]{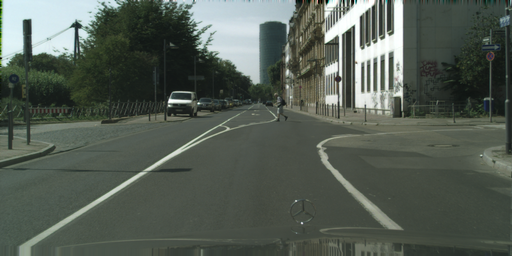}
\end{subfigure}
\begin{subfigure}{.5\textwidth}
	\includegraphics[scale=1.5, width=\textwidth]{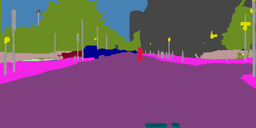}
\end{subfigure}
\begin{subfigure}{.5\textwidth}
	\includegraphics[scale=1.5, width=\textwidth]{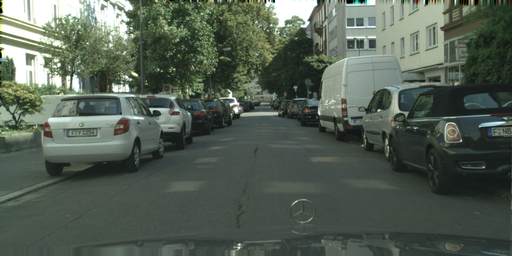}
\end{subfigure}
\begin{subfigure}{.5\textwidth}
	\includegraphics[scale=1.5, width=\textwidth]{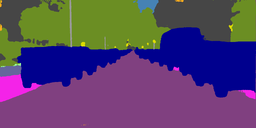}
\end{subfigure}
\begin{subfigure}{.5\textwidth}
	\includegraphics[scale=1.5, width=\textwidth]{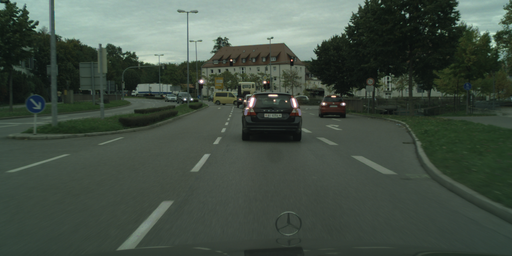}
\end{subfigure}
\begin{subfigure}{.5\textwidth}
	\includegraphics[scale=1.5, width=\textwidth]{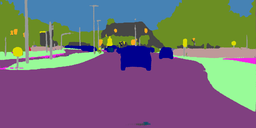}
\end{subfigure}

\caption{Qualitative examples of the Validation set and their segmented output from our model DG4*}
\vspace{-1em}
\end{figure*}

\section{Conclusion}
We approach the problem of designing extremely efficient CNNs for semantic segmentation specifically focusing on autonomous navigation. For this purpose we study the ERFNet model which is already among the most efficient, realtime models on the Cityscapes dataset. We first do a thorough benchmarking of ERFNet using the recent advances in efficient CNNs. We experiment with depthwise separable, grouped and shuffled convolutions. We observe that grouped convolutions along with depthwise separable convolutions can bring down the running time significantly, but results in over 10\% accuracy degradation. 

We propose a novel training procedure which can be easily implemented, which specifically targets grouped convolutions. The procedure starts with dense convolutions and gradually evolves toward grouped convolutions as the training progresses, allowing the optimization to be done at a higher dimensional space. We empirically show that this procedure on our proposed efficient architecture results in a model at run at 5.77 GFLOPs while giving reasonable accuracies.  

\bibliographystyle{ieee}
\bibliography{egbib}

\end{document}